\documentclass[runningheads]{llncs}
\usepackage{graphicx}
\usepackage{amsmath}
\usepackage{framed}
\begin{document}
\title{Diachronic Text Mining Investigation of Therapeutic Candidates for COVID-19}
\titlerunning{Diachronic Text Mining for COVID-19}
%
\author{James Powell\inst{1}\orcidID{0000-0002-3517-7485} \and
{Kari Sentz}\inst{1}\orcidID{0000-0002-1530-1952} }
\authorrunning{Powell et al.}
%
\institute{Los Alamos National Laboratory, Los Alamos, NM 87545, USA\\
\email{jepowell@lanl.gov}}
\maketitle              
\begin{abstract}
Diachronic text mining has frequently been applied to long-term linguistic surveys of word meaning and usage shifts over time. In this paper we apply short-term diachronic text mining to a rapidly growing corpus of scientific publications on COVID-19 captured in the CORD-19 dataset in order to identify co-occurrences and analyze the behavior of potential candidate treatments. We used a data set associated with a COVID-19 drug re-purposing study from Oak Ridge National Laboratory. This study identified existing candidate coronavirus treatments, including drugs and approved compounds, which had been analyzed and ranked according to their potential for blocking the ability of the SARS-COV-2 virus to invade human cells. We investigated the occurrence of these candidates in temporal instances of the CORD-19 corpus. We found that at least 25\% of the identified terms occurred in temporal instances of the corpus to the extent that their frequency and contextual dynamics could be evaluated. We identified three classes of behaviors: those where frequency and contextual shifts were small and positively correlated; those where there was no correlation between frequency and contextual changes; and those where there was a negative correlation between frequency and contextual shift. We speculate that the latter two patterns  are indicative that a target candidate therapeutics is undergoing active evaluation. The patterns we detected demonstrate the potential benefits of using diachronic text mining techniques with a large dynamic text corpus to track drug-repurposing activities across international clinical and laboratory settings.


\end{abstract}
\section{Introduction}
It seems increasingly unlikely that the virus that causes COVID-19 will be eliminated in the near-term. Some experts predict that COVID-19 will become a manageable endemic disease, but with the appearance of new variants with genetic mutations that increase transmission and cause higher rates of infection, along with the phenomena of fading immuno-response from vaccinations leading to breakthrough cases, it remains extremely important that safe, reliable therapeutics remain part of the overall strategy for managing this disease. Ongoing efforts in this regard include both the development of new treatments and the assessment of existing medications and small molecule compounds. There are many online biochemical libraries that publish metadata and provide a search interface for compound discovery. These compounds are typical evaluated in  in-vitro, in-silico, and in-vivo trials as appropriate. A fourth means of assessment can and should include text analysis of corpora containing scientific and clinical research related to these compounds. For COVID-19, this can include literature dating back to the emergence of SARS-COV-1 in 2003, as well as other coronaviruses that cause illness in humans. 

Diachronic word analysis is a Natural Language Processing (NLP) technique for characterizing the evolution of words over time. Often used in long-term historical linguistic studies, it can also be applied to scientific literature \cite{ref_Tshitoyan2019,ref_Dridi2019}, and has been shown to  reveal early evidence of scientific discoveries from large text corpora before they become widely known. 

In the 2019-2020 international global coronavirus pandemic of COVID-19, medical and pharmaceutical researchers around the world embarked on an ardent search for treatments, prophylactics, and cures. A part of that effort sought to identify existing drugs and therapeutics that might be useful in treating other diseases. Some of these re-purposing studies take advantage of the large amounts of data about drugs and infectious agents such as viruses and used computational approaches {\it{in-silico}}. Our study takes advantage of the constantly growing text corpora relevant to COVID-19 to understand how references to candidate treatments are evolving over time. It uses diachronic text mining with the goal of providing researchers with NLP tools to summarize the international published and pre-published trends.

The text corpus we used is the CORD-19 corpus \cite{ref_Wang2020b} that was established in March 2020 as a global repository for scientific and clinical research related to SARS-COV-2 and other coronaviruses. It aggregates content from PubMed, bioRxiv, medRxiv, and other sources and is updated with new publications on a regular basis. Figure 1 
illustrates the growth of CORD-19 through mid-2020.

\begin{figure}
\label{figure:cord-19}
\includegraphics[width=5.3in]{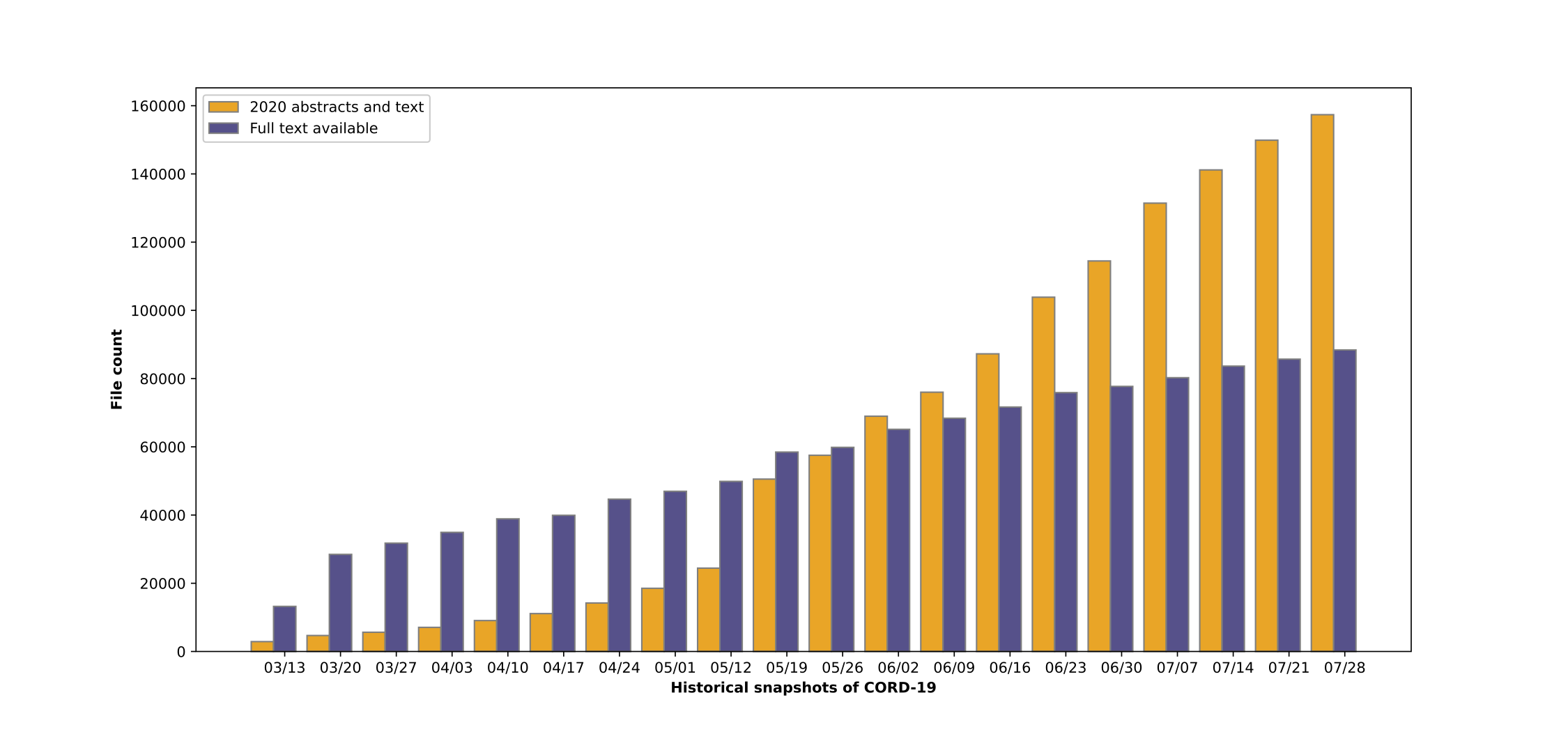}
\caption{CORD-19 Corpus Growth through July 2020}
\end{figure}

We used the CORD-19 corpus to conduct a diachronic text survey of candidate therapeutics identified in one of the more exhaustive drug re-purposing studies conducted for COVID-19, an in-silico study undertaken in February 2020 at Oak Ridge National Laboratory (ORNL). This study, detailed in a preprint published in Chemrxiv \cite{ref_Smith2020}, analyzed entries in the SWEETLEAD database for potential antiviral properties. SWEETLEAD is a  cheminformatics database of approved drugs, regulated chemicals, and herbal isolates. Over 9,000 of these small molecules were analyzed for their interactions with models of the SARS-COV-2 spike protein using a supercomputer at ORNL. The goal of the modeling effort was to determine which had theoretical potential to inhibit the virus's ability to infect human cells. They identified 41 candidates which modeling determined had significant theoretical efficacy. We used this list as targets for literature searches in the CORD-19 corpus. For comparison, we also analyzed a smaller list of candidate therapeutics published mid-summer 2020 in Nature \cite{ref_Shaffer2020}.
\newline
\newline
We summarize our study goals as follows:
\begin{enumerate}
\item{Given a target list of therapeutic candidates for treating COVID-19, how many can be detected in the CORD-19 corpus?}
\item{Do the semantic contexts and the candidate frequencies change in divergent or correlated ways over time?}
\item{Is this change meaningful - i.e. does it indicate heightened interest in a candidate therapeutic?}
\end{enumerate}

\section{Related Work}
Since digital corpora have become available, researchers have investigated various techniques for evaluating the long-term changes associated with the meanings or senses of words in large corpora over long time spans.\cite{ref_Kutuzov2018} They have found that raw frequency metrics have limited utility for this task. Some researchers have proposed that there are predictable behaviors related to frequency metrics verses metrics of semantic change. In \cite{ref_Hamilton2016a}, the authors introduced two “quantitative laws of semantic change” relating statistical word frequency to semantic change over time. Hamilton et al's law of conformity states that the rates of semantic change scale with the negative power of word frequency. In this investigation, we were guided by the observation that the combination of frequency and contextual metrics has yielded novel insights into how language changes over time.

Diachronic analysis of word characteristics in a corpus involves generating timeseries data for a given metric. Multiple studies demonstrated that word frequency metrics are useful for detecting sudden shifts in word occurrences over time including \cite{ref_Gulordava2011} and \cite{ref_Turney2010}.  Other studies used the Term Frequency-Inverse Document Frequency (TF-IDF) metric, which evaluates term frequency relative to a specific document as well as a document collection. The utility of this metric has been demonstrated in both search engines and machine learning models. 

More recently, studies have used various distributional representations of word contexts to identify semantic shifts such as those associated with two words with related meanings (polysemy) including \cite{ref_DelTredici2018}, \cite{ref_Martinc2019}. Still other studies have combined, compared, or contrasted these metrics \cite{ref_Gulordava2011}, using them to identify rules of semantic change \cite{ref_Hamilton2016a}. Only a few studies have utilized this sort of analysis to identify short-term linguistic changes in corpora spanning weeks or months instead of years, as in Stewart et al 2017 \cite{ref_Stewart2017} which examined frequency and linguistic shifts in social media posts. Even less common to date are investigations using these techniques to evaluate trends in scientific corpora.  One interesting example comes from Tshitoyan et al. \cite{ref_Tshitoyan2019}, where the authors' retroactively predict materials discovery in a historical corpus of materials science related abstracts using distributional approaches to latent semantic knowledge. 

Unsurprisingly, there is also a growing body of work related to text mining and COVID-19. Pozi et al \cite{ref_Pozi2020} demonstrated how word embeddings, t-sne  visualizations and k-means clustering could be used to summarize concepts in the  CORD-19 corpus.  Martin et al \cite{ref_Martinc2020} use Literature Based Discovery (LBD) techniques to mine scientific literature in order to discover bridges between genes and proteins related to COVID-19 using FastText and SciBERT contextual embeddings.  Parke et al \cite{ref_Park2020} devised a document clustering experiment with a small corpus of 34 papers related to drug repurposing and COVID-19 to discover where the literature agreed with respect to individual therapeutic candidates. Meanwhile, Tworowski et al \cite{ref_Tworowski2021} developed the COVID19 Drug Repository using text mining techniques to build a structured pharmacological dictionary of candidate COVID-19 therapeutics.  Wang et al \cite{ref_Wang2020b} developed EVIDENCEMINER, which is a tool that supports natural language queries for discovery of sentence level results for named entity queries from pre-indexed scientific corpora. Ahamed et al \cite{ref_Ahamed2020} constructed a network model of over 10,000 abstracts to mine information on transmission, drug types, and genome research related to coronaviruses. And in a similar vein, Rao et al \cite{ref_Rao2020} mined various text resources including abstracts from Medline and the CORD-19 to construct a network model focused on drug associations.


Our contribution is to use frequency and distributional text models to perform short-term diachronic text mining of a rapidly changing scientific corpus. Our goal was to characterize the behavior of a target list of candidate therapeutics for COVID-19 over a relatively short period of time. More specifically, as our investigation was concerned with short-term changes in word occurrence and usage patterns, we focused on a document-oriented frequency metric (TD-IDF) and a distributional metric (cosine) based on temporal word embeddings. \cite{ref_JurafskyMartin2020} 
These metrics were used to calculate timeseries data for named candidate therapeutics appearing in historical snapshots of the CORD-19 corpus. Our hypothesis was that analysis of these two streams of data would allow us identify small but potentially significant patterns of change associated with these candidates that might not otherwise be apparent.

\section{Data Sources} 

The CORD-19 corpus was established by Microsoft in March, 2020 to serve as a repository of peer-reviewed papers and preprints related to the SARS-COV-2 virus, other coronaviruses, and COVID-19. This repository was updated weekly until late May, at which time the maintainers began to update it daily. We used weekly historical snapshots of the corpus since this allowed enough time to pass for a significant number of changes to accumulate, which is supported by document counts in Figure 1. We limited our survey to  snapshots spanning March 13 until July 28, 2020. During that time, the CORD-19 corpus expanded from 13,219 full text documents to 88,399 papers by period end. We included only full papers in our analysis. 

Along with historical snapshots of the CORD-19 corpus, we used a dataset produced by the ORNL study as our source for candidate therapeutic search targets. This data set, which listed 9,128 candidates, had several advantages: 1) it was one of the earliest COVID-focused repurposing studies at the time of its publication (late March 2020), 2) it was an exhaustive study, characterizing thousands of compounds, 3) it included searchable names for the candidates, 4) it was published in ChemRxiv, a preprint  archive not included in CORD-19. Since March, there have been many other drug repurposing studies. So for completeness, we decided to  analyze a second smaller list of potentially repurposable therapeutics for COVID-19 that appeared in Nature, in May 2020.\cite{ref_Shaffer2020}   




\section{Methods}

For text pre-processing, we used a hybrid phrase detection algorithm based employing a dictionary as described in Lui et al \cite{ref_Liu2020} and based on an approach described in \cite{ref_Mikolov2013}. First, we seeded the co-occurrence detection algorithm with entries from the candidate therapeutics list such as``folic acid'' and ``fluticasone propionate'', since these precise strings are already known. Then it applied a count-based equation to identify additional phrases. We normalized each phrase to a form where spaces were replaced with an underscore. This approach increased the likelihood that concept phrases such as ``monoclonal antibody'' and ``recommended dosage'' would be found. 

For analysis, we use the well-known frequency metric TF-IDF. \cite{ref_JurafskyMartin2020} This metric captures term frequency in a corpus proportional to inverse document frequency, resulting in a score that measures the relative importance of a term in a corpus. This particular metric has been used in many information retrieval tasks because it is able to identify significant terms across a large corpus. 
In addition to frequency metrics, we utilized distributional methods (word embeddings) to model semantics of terms in the corpus. Word embeddings are an unsupervised machine learning technique used to model word meanings in a corpus, based on word co-occurrence patterns. It reveals clusters of words that have similar semantic or syntactic patterns. There are a number of methods that emerge from the simple idea that the meaning of a word is defined in some way by how often it occurs near other words. \cite{ref_JurafskyMartin2020} This evokes Firth's famous 1957 quote ``You shall know a word by the company it keeps!'' as quoted in \cite{ref_JurafskyMartin2020}. These ``word neighborhood'' methods represent a word by vectors of words that co-occur with it and these can be created based on large corpora to gain insight into semantics through statistics. These vectors are called {\it{embeddings}}. We utilize a particular version of word embeddings to examine the CORD-19 corpus across time and analyze representational changes from instance to instance. These {\it{Temporal Word Embeddings with a Compass}} (TWEC) word embeddings were developed by Di Carlo et al. in their 2019 publication.\cite{ref_DiCarlo2019} 
TWEC builds an embedding model based on the aggregation of all instances of the corpus and then generates an embedding model for each historical snapshot of the corpus.

Each metric was used to model the frequency and semantic behavior of each target term per historical snapshot of the CORD-19 corpus. To evaluate how the metrics for each term change over time, we accumulate these values into temporal vectors per target term, per metric. Given a target term X, we generate two temporal vectors Y and Z, where Y contains TF-IDF scores per corpus temporal snapshot, and Z contains cosine distance to an atemporal word embedding model from each temporal embedding snapshot. The Y vector is used to calculate change values in the frequency metric, resulting in a Y' vector containing deltas between historical snapshots. These two vectors are then used to analyze term behavior within the corpus over time, and to find other terms not in the target list that are most similar in some fashion, based on these temporal vectors. We also look for per term correlations in temporal patterns that might indicate that the term's presence and meaning within the corpus was volatile, stable, or exhibited a similar change pattern in both frequency and semantic shift. For the TF-IDF metric, these calculations are straightforward, and the resultant delta vector is a floating point value that represents the the amount of change, that is, the difference between two adjacent TF-IDF scores in time. Analyzing semantic shift requires generating diachronic word embeddings which are distributional word embedding models for each temporal instance of the corpus, but also grounded in an atemporal model that is learned over all documents from all temporal instances of the corpus. This allows meaningful comparisons to be made between the word embedding models, such as measuring the cosine distance for a target between temporal instances, or between a temporal instance and the atemporal model. We used a temporal word embedding implementation (TWEC) described in \cite{ref_DiCarlo2019}. The central part of the model is the updating of the input matrix for each slice and the temporal model interpretation. The authors use a negative sampling \cite{ref_Mikolov2013} to calculate the softmax function $\sigma$ to maximize the probability $P$ that a set of words surrounding a target word, $w_k$, are representative of its context in time ($C^t$), when multiplied by the mean of atemporal word embedding vectors from $u$ (the compass) for the same set of context words around $w_k$ for any given temporal slice. This is given by the optimization equation:

\begin{equation}
  \mathnormal {\max_{ \mathbf{C}^t }\log P({w_k}|\gamma({w_k})) = \sigma (\vec{u}_k \cdot \vec{c} {\,^t_{\gamma(w_k)}})} 
  \label{eq:TWEC}
\end{equation}

\noindent where  $\vec{c} {\,^t_{\gamma(w_k)}}$ are context embeddings for word $w_k$, $\vec{u}_k$ are the target embeddings, and $\gamma({w_k})$ represents the $M$ words in the context of $w_k$. The mean of the temporal context embeddings, $\vec{c} {\,^t_{j{_m}}}$ of the context words, $w_{j{_m}}$, is given by the equation, \cite{ref_DiCarlo2019}: 

\begin{equation}
     \vec{c} {\,^t_{\gamma(w_k)}}=\frac{1}{M}(\vec{c} {\,^t_{j_1}}+...+\vec{c} {\,^t_{j_M})^T}
\end{equation}

We trained the word embedding model using the continuous bag of words (CBOW) architecture for the underlying  shallow neural network with the following parameters: a window size of 5 (the context around each word), a minimum word occurrence of 3, and a vector length of 100. These parameters determine the quality of the resulting word embeddings. For each temporal instance, the network learns a context vector for any word or pre-encoded phrase that occurred more than 3 times, applying the objective described in Equation \ref{eq:TWEC}. After processing all weekly historical snapshots of the corpus, we ended up with 21 word embedding models (20 weekly models plus the atemporal compass embedding model).

To measure semantic shift, we assemble a vector of cosine distance values for temporal instance, for each candidate therapeutic. Each value is computed with the cosine distance metric, between the context of the candidate $C$ at time $t$, $C^t$. This represents the extent to which the term's context has changed over the time period. We also captured nearest neighbors for each candidate from each historical snapshot, also using the cosine metric. A visualization of this can be see in Figure 2. 

\noindent
\begin{figure}
\label{Figure:sankeys}
\begin{tabular}{|c|c|}
\hline
\includegraphics[width=2.5in]{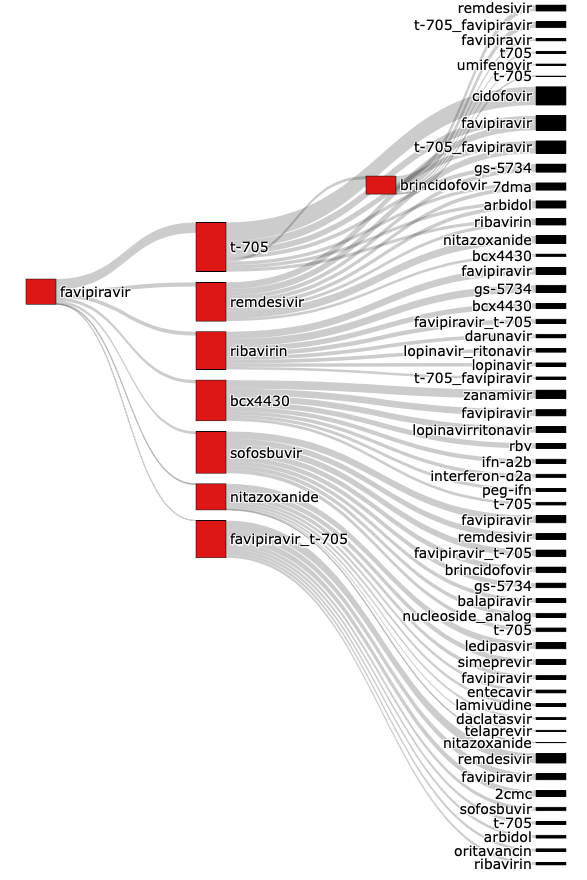}
&
\includegraphics[width=2.5in]{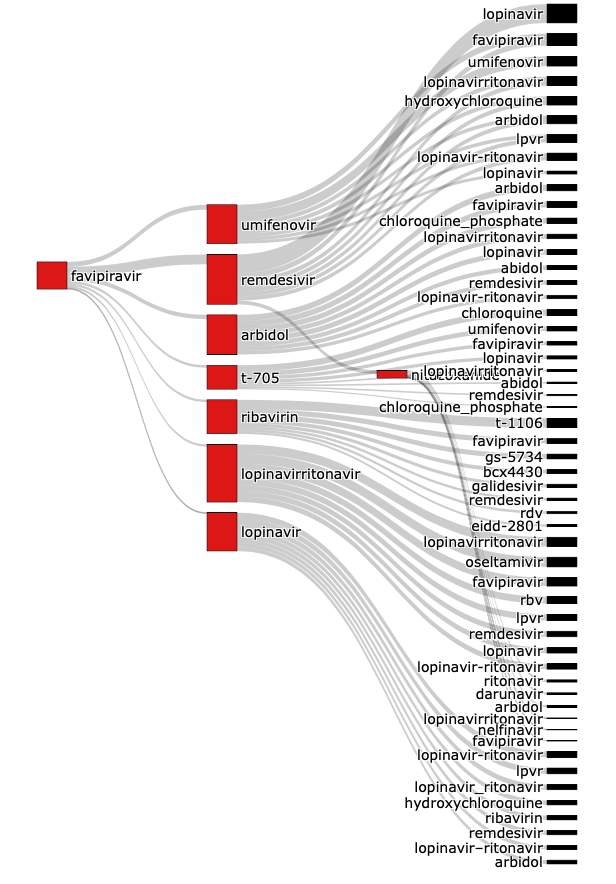}\\
\hline
\end{tabular}
\caption{How the relationship as reflected in historical word embedding models for candidate {\it{favipiravir}} changed from 3/13 to 7/28}
\end{figure}

\section{Results}
We performed a search in each historical snapshot of the CORD-19 corpus for each candidate therapeutic, using the normalized form of the name as described above. We detected 14\% (1267) of the candidate therapeutics in CORD-19 on March 13, 2020. This increased to 26\% (2388) by July 28, 2020. We evaluated the frequency and context of candidates found in four or more adjacent historical snapshots of the corpus.

\begin{figure}
\begin{center}
\label{Figure:corr-boxplot}
\includegraphics[width=3.75in]{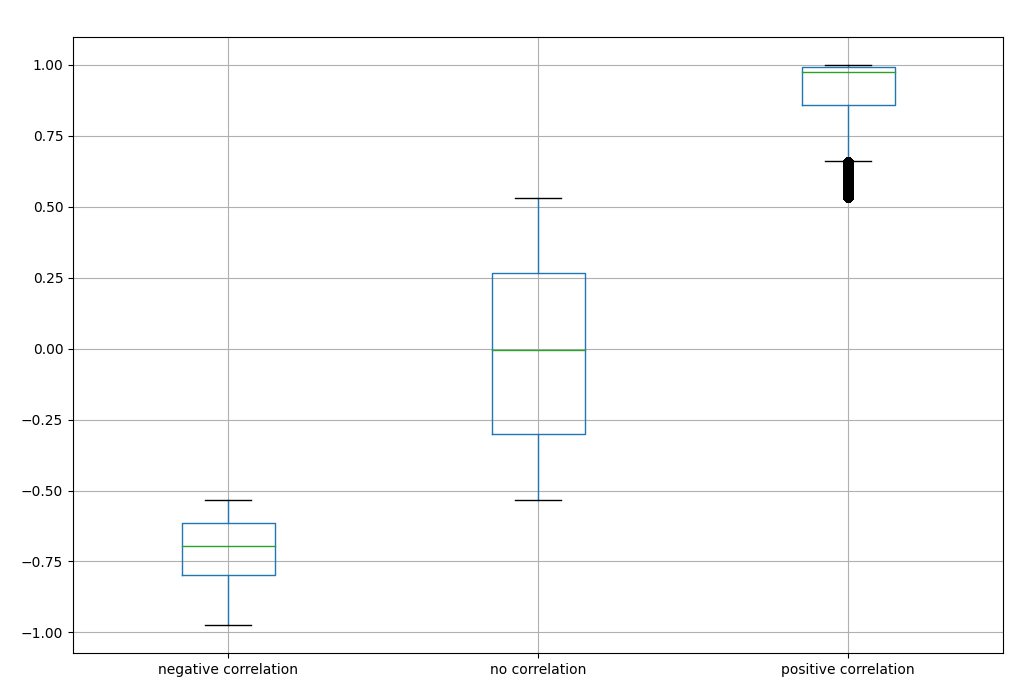}
\end{center}
\caption{Correlation coefficient plot for TF-IDF vs cosine metric timeseries data for detected candidates from the Oak Ridge study. Each plot reflects the distribution of correlation scores in a band.}
\end{figure}

In the Oak Ridge drug repurposing study preprint, the authors concluded their paper with a list of seven candidates that they considered worthy of further investigation.  Here we summarize our findings related to five of these seven candidates in Table 1. The other two candidates, {\it{pemirolast}} and {\it{ergoloid}} did not appear in the CORD-19 corpus through July 28.

\begin{center}
\small
\begin{table}
\caption{Oak Ridge Study Candidates : frequency and correlation}
\label{Table:ORNLCandidates}
\begin{tabular}{|l|l|l|l|l|l|l|}
\hline
\multicolumn{1}{|p{2cm}|}{\textbf{Drug Name}} & \multicolumn{1}{|p{2cm}|}{\textbf{Type}} &
\multicolumn{1}{|p{1.5cm}|}{\textbf{Frequency at Start}} &
\multicolumn{1}{|p{1.5cm}|}{\textbf{Frequency at End}} &
\multicolumn{1}{|p{1.5cm}|}{\textbf{Cosine Distance}} &
\multicolumn{1}{|p{1.5cm}|}{\textbf{TF-IDF Score}} & 
\multicolumn{1}{|p{2cm}|}{\textbf{Correlation Coefficient}} \\
\hline

\hline
cepharanthine&anti-inflammatory&7&126&0.72&12.53&-0.18\\
\hline
eriodictyol&herbal isolate&2&27&0.72&13.27&0.14\\
\hline
hypericin&herbal isolate&26&263&0.68&11.49&0.11\\
\hline
isoniazid&antibiotic&73&1149&0.75&9.78&-0.09\\
\hline
nitrofurantoin&antibiotic&9&390&0.72&10.93&-0.26\\
\hline
\end{tabular}
\end{table}
\end{center}

For those candidates that met the selection criteria, we were able to calculate TF-IDF scores and measure semantic change as described above. This set of frequency and context measures from all of the historical snapshots constituted the respective TF-IDF and semantic change timeseries vectors for each candidate. Lower TF-IDF scores are ``better'' in that they indicate that a word is relatively uncommon across the corpus. Cosine is a measure of the angle between two vectors, so a higher cosine value means that the angle is smaller, and thus the terms being compared are closer to one another. 

\begin{table}
\caption{Oak Ridge Study : Similarity Matrix Best Matches}
\label{Table:ORNLCandidatesSimilarities}
\begin{tabular}{|l|l|l|}
\hline
{\bf{Name}} &{\bf{Most similar TF-IDF timeseries}}&{\bf{Most similar cosine timeseries}}\\
\hline
\hline
cepharanthine&finasteride&zopiclone\\
\hline
eriodictyol&taxifolin&pancuronium\\
\hline
hypericin&glycerine&piperazine\\
\hline
nitrofurantoin&penicillamine&testosterone\\
\hline
isoniazid&tenofovir&ethambutol\\
\hline
\end{tabular}
\end{table}

To analyze the relationship between the timeseries vectors for TF-IDF and embedding cosine distance for each candidate therapeutic, we calculated the Pearson correlation coefficient between these two vectors, where values for $x$ are TF-IDF scores for each historical snapshot of the corpus, and values for $y$ are cosine distance for each snapshot:

\begin{equation}
  r =
  \frac{ \sum_{i=1}^{n}(x_i-\bar{x})(y_i-\bar{y}) }{%
        \sqrt{\sum_{i=1}^{n}(x_i-\bar{x})^2}\sqrt{\sum_{i=1}^{n}(y_i-\bar{y})^2}}
\end{equation}
\noindent The value of $r$ associated with a candidate therapeutic indicates where there is a positive, negative, or no correlation between the frequency and cosine distance timeseries vectors. 

We used this same approach to evaluate a second list of fifteen drug candidates identified in a peer-reviewed paper in \cite{ref_Shaffer2020}. Of the candidates named in this paper, three were monoclonal antibodies, so we omitted these as our experimental setup was focused on chemical compounds. We also could not directly search for the fourth candidate as it was a class of compounds (steroids). However, we did include {\it{dexamethasone}} as a representative for that class. We were able to analyze a total of 12 entries from this list summarized in Tables 3 and 4. In Figure 4, we see dramatic differences between the stability of {\it{dexamethasone}} over time with consistent colors across the horizontal axis in comparison to {\it{nitrofurantoin}} that changes dramatically one time step to the next.

\begin{center}
\small
\begin{table}
\caption{Nature article candidates}
\label{Table:NatureCandidates}
\begin{tabular}{|l|l|l|l|l|l|l|}
\hline
\multicolumn{1}{|p{2cm}|}{\textbf{Drug Name}} & \multicolumn{1}{|p{2cm}|}{\textbf{Type}} &
\multicolumn{1}{|p{1.5cm}|}{\textbf{Frequency at Start}} &
\multicolumn{1}{|p{1.5cm}|}{\textbf{Frequency at End}} &
\multicolumn{1}{|p{1.5cm}|}{\textbf{Cosine Distance}} &
\multicolumn{1}{|p{1.5cm}|}{\textbf{TF-IDF Score}} & 
\multicolumn{1}{|p{2cm}|}{\textbf{Correlation Coefficient}} \\
\hline
\hline
camostat&protease inhibitor&37&794&0.69&10.77&-0.66\\
\hline
chloroquine&anti-malarial&900&18210&0.77&7.74&0.87\\
\hline
dexamethasone&steroid&288&5201&0.76&8.46&0.69 \\
\hline
famotidine&histamine-2 blocker&18&666&0.69&11.05&-0.57\\
\hline
fluvoxamine&antidepressant&1&58&0.72&13.05&-0.29\\
\hline
hydroxychloroquine&anti-malarial&48&16623&0.75&8.62&-0.41\\
\hline
ivermectin&anti-parasitic&47&2638&0.76&9.39&-0.02\\
\hline
lopinavir&anti-viral&104&3589&0.78&8.98&0.83\\
\hline
nafamostat&protease inhibitor&12&423&0.65&11.72&-0.04\\
\hline
nitazoxanide&anti-parasitic&144&1788&0.75&9.69&-0.82\\
\hline
ritonavir&antiviral&99&3114&0.74&8.96&0.72\\
\hline
\end{tabular}
\end{table}
\end{center}

\begin{figure}
\label{Figure:heatmaps}
\begin{tabular}{|c|c|}
\hline
\includegraphics[width=2.7in]{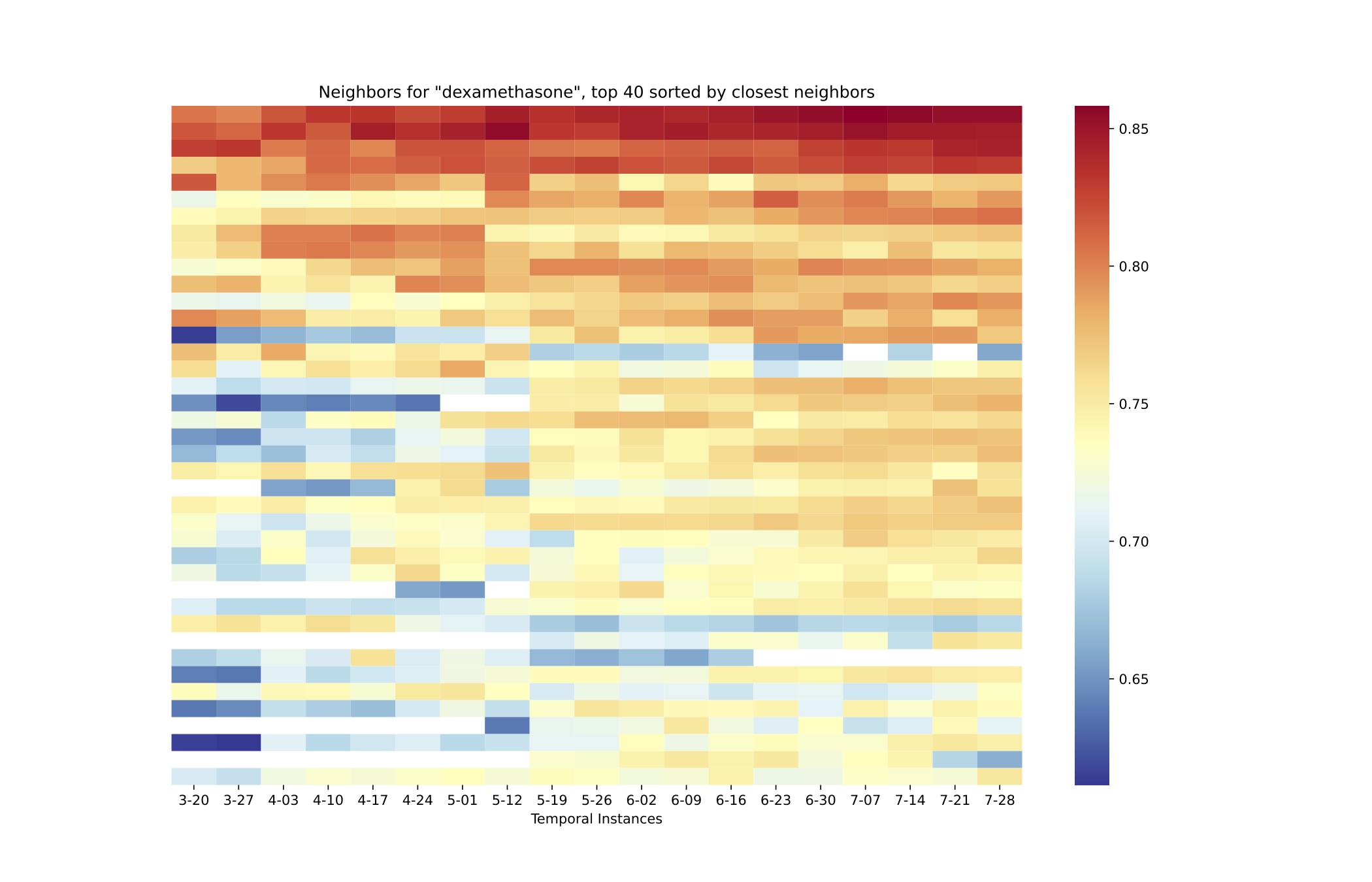}
&
\includegraphics[width=2.7in]{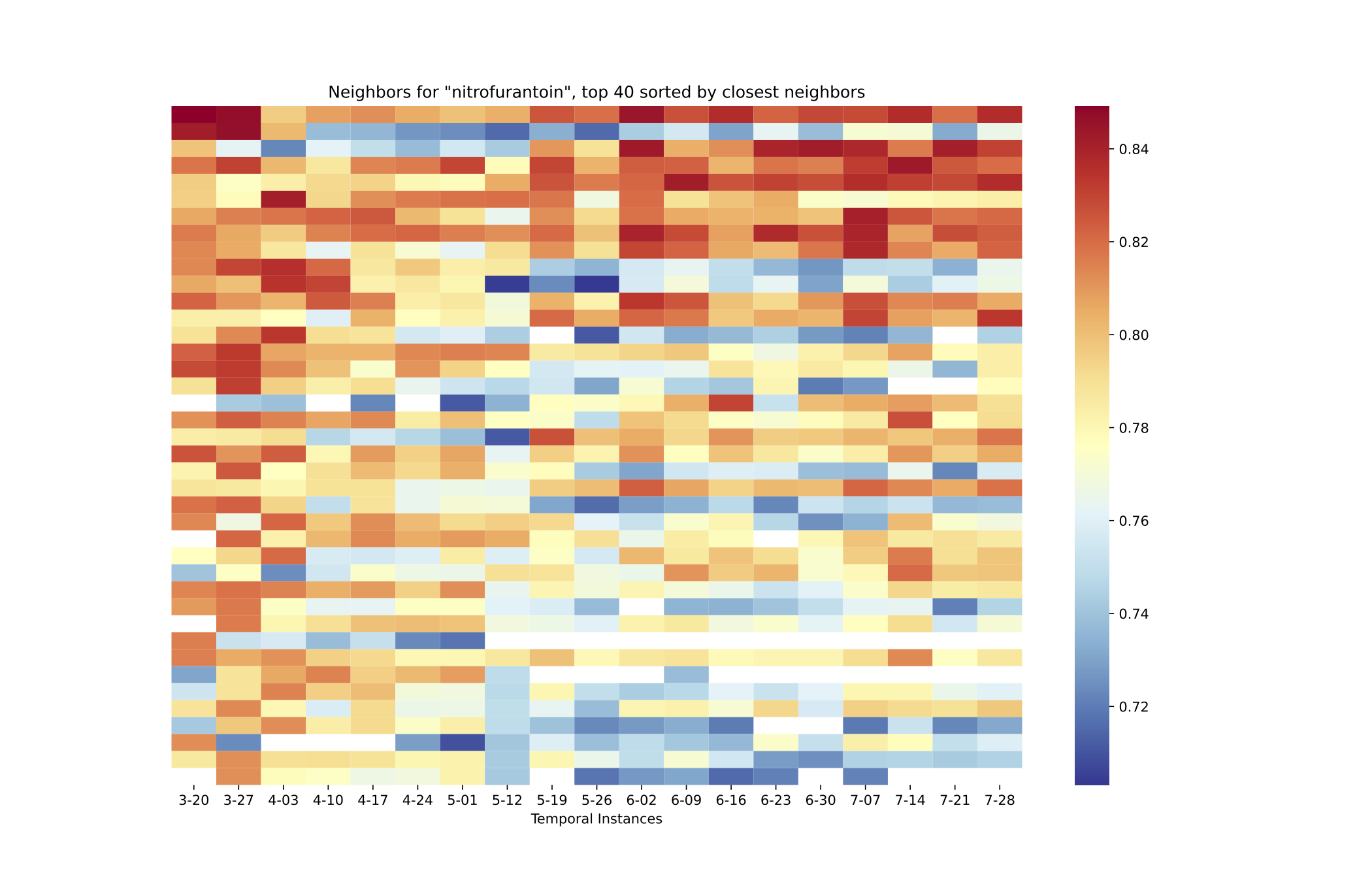}\\
\hline
\end{tabular}
\caption{On the left, a heatmap illustration of the changing relationship of {\it{dexamethasone}} to its neighbors in the embedding models over time. Note that it is relatively stable, while an illustration for another candidate {\it{nitrofurantoin}} on the right shows considerable change over time. }
\end{figure}

\begin{table}
\caption{Nature candidates : Similarity Matrix Best Matches}
\label{Table:NatureCandidatesSimilarities}
\begin{tabular}{|l|l|l|}
\hline
{\bf{Name}} &{\bf{Most similar TF-IDF timeseries}}&{\bf{Most similar cosine timeseries}}\\
\hline
\hline
camostat&catechol&lycorine\\
\hline
chloroquine&glycerol&dexamethasone\\
\hline
dexamethasone&acyclovir&chloroquine\\
\hline
famotidine&cefixime&glucuronic acid\\
\hline
fluvoxamine&iomeprol&fingolimod\\
\hline
hydroxychloroquine&lopinavir&ibuprofen\\
\hline
ivermectin&sulfamethoxazole&vorinostat\\
\hline
lopinavir&ritonavir&dexamethasone\\
\hline
nafamostat&ephedrine&octanol\\
\hline
nitazoxanide&artemisinin&niclosamide\\
\hline
ritonavir&lopinavir&alcohol\\
\hline
\end{tabular}
\end{table}

\begin{figure}
\includegraphics[width=\textwidth]{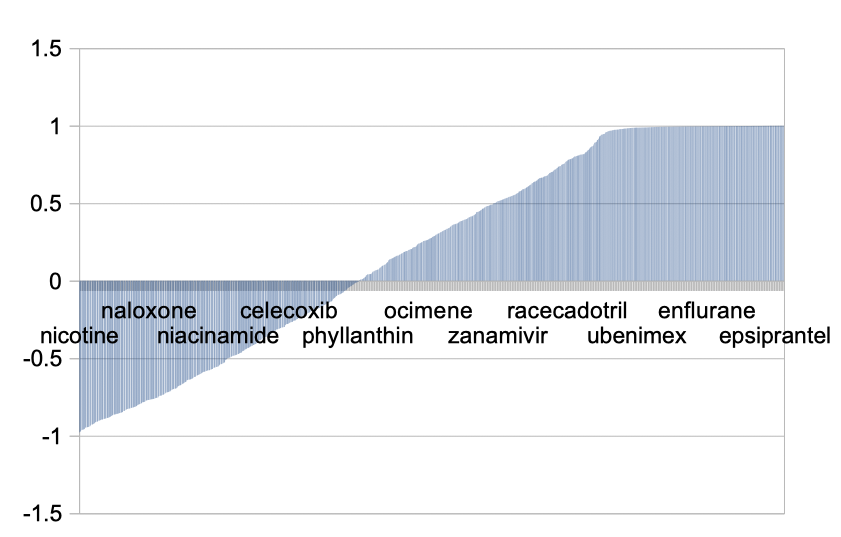}
\caption{Correlation coefficient plot for TF/IDF vs embedding for detected candidates}
\label{fig4}
\end{figure}

 \section{Discussion}
Therapeutic candidate detection by name was successful, with 26\% of small molecule candidates from the Oak Ridge dataset identified, and detection of 80\% of the candidate therapeutics from  {\it{Nature}} \cite{ref_Shaffer2020}. Among the detected candidate therapeutics, we were able to identify three distinct categories of behavior based on correlation between frequency and semantic change metrics: positive (>.53 correlation coefficient), negative (-.53 < correlation coefficient), and those with correlation coefficients falling between -.53 and .53. Around 45\% of detected candidates exhibited a positive correlation between their frequence and semantic change data, that is, their frequency and semantic change rates were stable, increasing, or decreasing at a similar rate.  When a candidate did exhibit a positive correlation between its TD-IDF and TWEC in the timeseries data, we observed that it tended to be a drug that exhibited a high frequency of occurrence within the corpus, starting with very early historical snapshots of the corpus (compare semantic change heatmaps in Figure 4). In other words, they were already known entities, and thus not likely to exhibit significant changes or divergence over time. About 14\% of the detected candidates exhibited a negative, or inverse correlation between their frequency and semantic change rates. Finally, about 40\% of the detected candidates exhibited no significant correlation between the rate of change for these two metrics. 

It appears that the most useful insights result from comparison of the accumulated timeseries data per candidate. The timeseries data is simply a vector of the frequency based TF-IDF score and cosine distance measure for each week that a named candidate appears in the corpus. Candidates with uncorrelated or inverse correlated TF-IDF and semantic change timeseries data tend to correspond to viable candidate treatments, including antibiotic, antiparasitics and antivirals. Evaluating the cosine similarity among the accumulated TF-IDF timeseries for each candidate performing the same analysis using the cosine distance   semantic change timeseries values reinforces expected relationships among the candidates. Items with similar timeseries tend to have meaningful biological similarity as well. Interestingly, TF-IDF seemed to edge out semantic change timeseries in terms of appropriateness of associations (Table \ref{Table:ORNLCandidatesSimilarities} and Table \ref{Table:NatureCandidatesSimilarities}). It suggests that comparison of timeseries data for candidate therapeutics could yield increasingly useful insights as the corpus grows, even if this data does not yield useful insights week over week.

This study outlines techniques for characterizing and analyzing the occurrences of named candidate therapeutics in multiple temporal snapshots of a relevant corpus using TF-IDF frequency and distributional semantic timeseries data. It suggests that analyzing these metrics independently and in combination over time might be used to infer the potential efficacy or contrary indicators to efficacy for candidates that occur in the corpus.  The value of these two metrics is reinforced when computing a similarity matrix over all of the candidates, using each metric independently. This may mean that, given sufficient data, future trends in these metrics could be forecast using a timeseries modeling and forecasting techniques such as ARIMA. When a forecast of future behavior for a candidate therapeutic aligns with the observed behavior of a drug with proven efficacy that also occurs in the same corpus, the forecast candidate may be worthy of further scrutiny.

\begin{figure}
\begin{center}
\label{Figure:traj=acetazolamide}
\includegraphics[width=3.75in]{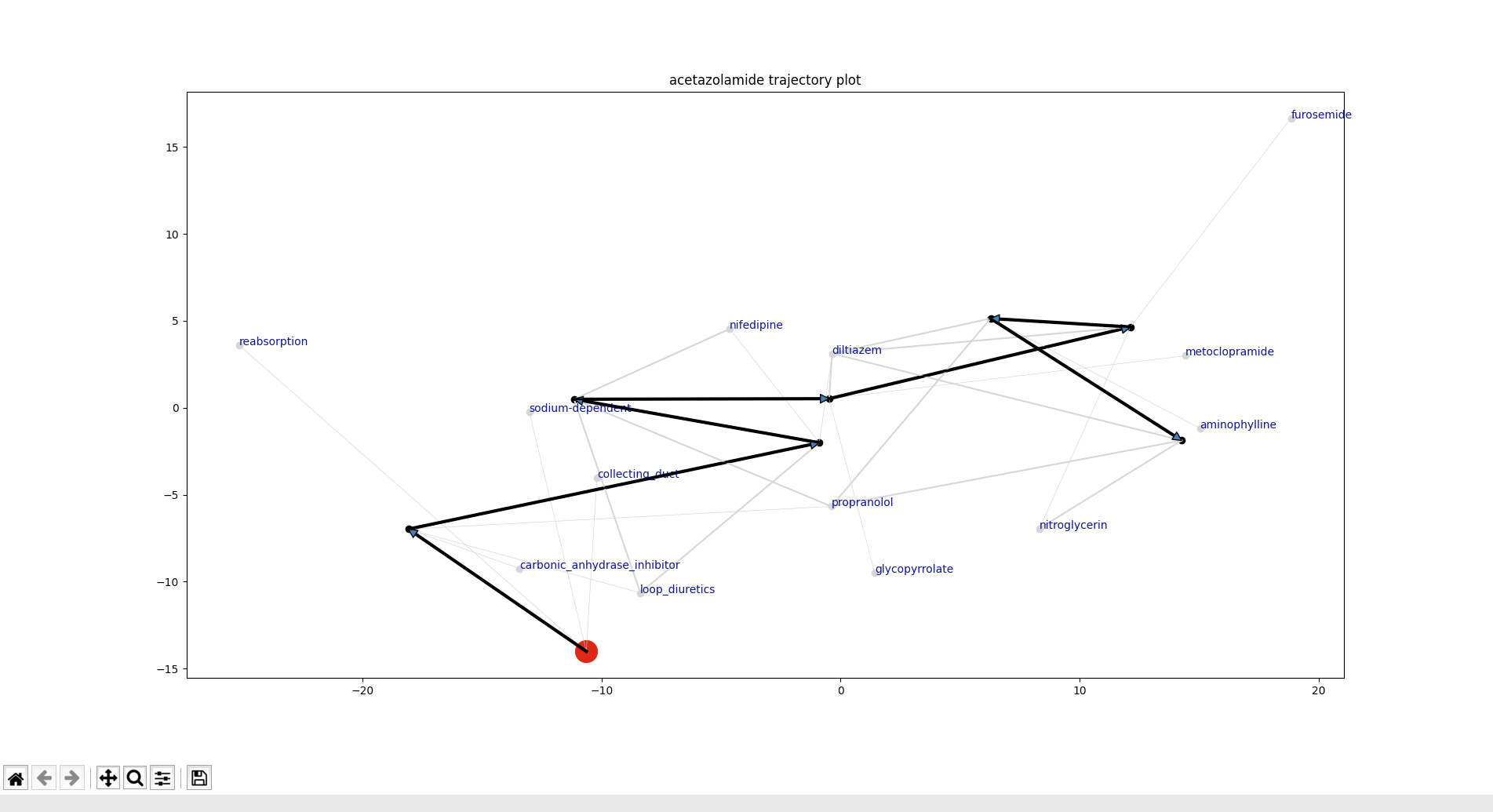}
\end{center}
\caption{t-SNE trajectory of acetazolamide through the temporal embedding space: March-July 2020}
\end{figure}

It is not currently possible to conduct an objective, quantitative  evaluation of how well our metrics performed, since there is not yet a ground truth list of candidate therapeutics that have been successfully repurposed for COVID-19. Unfortunately this is a common problem. Once one or a few treatments are developed for a disease, there's little incentive to continue to investigate other drugs for  repurposing.  We can comment on some limitations we observed. There were notable instances that revealed how both frequency and embedding models can fall short due to data sparsity, overfitting, false equivalencies, and a phenomenon called ``leaked training data''. This was revealed when we attempted to match drugs based on their TF-IDF and embedding distance timeseries vectors, again using the cosine metric to find the most similar timeseries vectors. Problems include:
 
\begin{itemize}
\item{Sparse data resulting in outlier clusters with poor semantic similarity such as {\it{ibuprofen}} and {\it{hydroxychloroquine}} in Table 4. }
\item{Overfitting occurs when too many similar examples drown out the effects of rarer contexts, as illustrated in Table 4 for the cosine entry for {\it{ritonavir}} with  {\it{alcohol)}}. }
 \item{False equivalences identified with frequency metrics, such as the TF-IDF match between {\it{hypericin}} and {\it{glycerine}} in Table 2. }
 \item{Leaked training data resulting in what appeared to be  meaningless contextual associations. For example, {\it{indometacin}} was initially most closely associated with strings of numbers, including 68.1, 41.0, 46.0. Upon further investigation we traced this back to a  pre-COVID clinical trial study in the earliest CORD-19 corpus snapshot. Later {\it{indometacin}} had moved closer to anti-inflammatory and pain medications.  }
 \end{itemize}
 
\noindent We think it is important to keep these examples in mind, even though they were uncommon, as they expose some of the limits of these text mining techniques. Short-term diachronic text analysis appear to be highly sensitive to these issues.

\section{Conclusion and Future Work}
In our short-term diachronic survey of candidate therapeutics in the CORD-19 corpus, we demonstrated that we could detect candidate therapeutics, and measure changes of word occurrences and changes to word contexts over time. While 45\% exhibited trivial changes characterized by highly correlated semantic and frequency shifts over time, 55\% exhibited changes where frequency and semantic shift patterns diverged (Figure 3). 
We believe these reflect event-driven shifts associated with these candidates that may correspond to new research findings or clinical determinations of efficacy. We could envision the combination of candidate detection, measurement, and finer-grained visualization techniques forming the core of a text mining tool that provides a novel means of exploring a dynamic scientific corpus over time. 

In the future we plan to expand the number of historical snapshots in our survey, extract concordance data (keywords in context) to better understand how candidates are changing, and investigate how to provide the capabilities we've outlined in this paper as an interactive tool. In addition, a drug name database will be incorporated to detect name variants, brand names, etc. Finally we plan to investigate the possibility of using language models to classify the deeper semantics associated with the observed changes.

%
%
%

\end{document}